\documentclass{article}
\usepackage{ijcai26}

\usepackage{times}
\usepackage{soul}
\usepackage{url}
\usepackage[hidelinks]{hyperref}
\usepackage[utf8]{inputenc}
\usepackage[small]{caption}
\usepackage{graphicx}
\usepackage{amsmath}
\usepackage{amsthm}
\usepackage{enumitem}
\usepackage{xspace}
\newcommand{\eg}{e.g.\xspace}
\usepackage{amssymb}
\usepackage{booktabs}
\usepackage[all]{hypcap}
\usepackage{algorithm}
\usepackage{algorithmic}
\usepackage{multirow}
\usepackage[switch]{lineno}

\usepackage{latexsym}

\title{CMAP: Cross-Modal Adaptive Prompting for Multi-Domain Task-Incremental Learning}

\author{
    Sriram Mandalika
    \affiliations
    Hasso Plattner Institute
    \emails
    Potsdam, Germany
}

\begin{document}

\maketitle

\begin{abstract}
Multi-domain task-incremental learning requires a model to sequentially acquire knowledge across visually diverse domains without forgetting prior tasks, and without access to task identity at inference. Parameter-efficient methods built on frozen vision-language models have made strong progress, yet all existing approaches rely exclusively on visual features for task routing, confidence estimation, and encoder adaptation, leaving CLIP's cross-modal text embedding space entirely unexploited. We address this gap through three contributions. Text-space task routing replaces visual Gaussian matching with cosine similarity to frozen CLIP text prototypes, giving order-independent routing robust to data scarcity at zero parameter cost. Multi-prototype visual-textual confidence replaces single-Gaussian class modeling with K-means visual prototypes and cross-modal alignment scores under task-calibrated thresholds. Symmetric cross-modal gating extends per-layer Gumbel gates to the text encoder conditioned on batch image features, preserving cross-modal alignment on out-of-distribution inputs. On the MTIL benchmark spanning 11 datasets and 1201 classes, our method achieves 74.2\% Transfer, 80.5\% Average, and 88.7\% Last under Order-I, surpassing the prior state of the art by 5.0, 3.7, and 3.0 percentage points with only 2.5M trainable parameters and no external data.
\end{abstract}

\section{Introduction}

Modern vision-language models such as CLIP~\cite{radford2021learning} have demonstrated remarkable zero-shot generalisation by learning rich semantic alignments between visual and textual representations from large-scale data. Deploying these models in real-world applications, however, often requires learning new tasks sequentially rather than
all at once. This setting, known as Multi-Domain Task-Incremental
Learning (MTIL)~\cite{zheng2023preventing}, presents two fundamental challenges. As a model adapts to incoming tasks, it tends to overwrite previously acquired knowledge, a phenomenon known as catastrophic forgetting~\cite{zhou2024continual}. At the same time, adapting too aggressively to seen tasks degrades the
model's ability to generalise to unseen ones, a problem referred to
as forward forgetting~\cite{zheng2023preventing}. Solving both simultaneously, without
access to past data and without knowing the task identity at
inference, remains an open and practically important problem.

Parameter-efficient fine-tuning has emerged as the dominant
paradigm for addressing MTIL with pre-trained vision-language
models. By keeping the backbone frozen and introducing a small
number of learnable parameters such as prompts or adapters, these
methods aim to adapt the model to new tasks while preserving its
pre-trained knowledge~\cite{Learingdomain,zhou2024continual}. The most recent state of the art,
IAP \cite{fu2026iap}, advances this line of work by introducing instance-aware gated
prompting, where each transformer layer decides whether to apply a
prompt based on the individual input, and a two-stage Gaussian
confidence cascade that weighs prompting intensity according to how
confidently an instance belongs to a known task distribution. These
are meaningful steps forward, but they share a fundamental
assumption that all prior MTIL methods make, that adaptation control
should be driven entirely by visual features.

This assumption overlooks a defining property of CLIP~\cite{radford2021learning}. Unlike
purely visual models, CLIP is trained to align images and text in a
shared embedding space, and its text encoder encodes rich,
semantically stable class knowledge that is entirely independent of
any visual training distribution. When a MTIL method selects which
task's prompts to apply using visual Gaussian statistics, it ignores
this cross-modal signal. When it estimates instance confidence using
only visual prototype distances, it ignores the image-text alignment
that directly determines classification accuracy. And when it applies
adaptive gating to the image encoder but leaves the text encoder on
fixed-depth prompting, it breaks the cross-modal coherence that CLIP
was trained to maintain. Existing methods treat a cross-modal model
as if it were a visual-only one, and this mismatch is a significant
source of untapped potential.

Motivated by the above observations, we propose \textbf{CMAP}: \textbf{C}ross-\textbf{M}odal \textbf{A}daptive \textbf{P}rompting,
a novel framework for MTIL that addresses the visual-only bias present
in all existing methods. CMAP extends IAP with cross-modal awareness
at every stage of inference. We replace visual Gaussian task routing
with cosine similarity to frozen CLIP text prototypes that are
computed once before training, require no additional learnable
parameters, and remain invariant to task arrival order and training
sample count. We introduce Multi-Prototype Visual-Textual Confidence,
which represents each class with multiple visual cluster centroids and
combines their similarity scores with cross-modal image-text alignment
for a more reliable confidence estimate, alongside task-adaptive
thresholds that replace the fixed global thresholds used in prior
work. We further extend per-layer Hard Gumbel gating symmetrically
to the text encoder, conditioned on image features, so that both
encoders co-adapt and co-suppress in response to each input,
preserving cross-modal alignment on out-of-distribution inputs.
Together, these contributions ensure that both modalities of CLIP
participate in adaptation decisions throughout the incremental
learning process.

The contributions of CMAP can be summarised as follows.

\begin{itemize}

\item We introduce text-space task routing, replacing visual Gaussian
routing with cosine similarity to frozen CLIP text prototypes. This
requires no additional learnable parameters and remains robust to
task arrival order and limited training data, making it especially
effective in data-scarce continual learning settings.

\item We propose Multi-Prototype Visual-Textual Confidence (MPVTC),
which models each class with multiple visual cluster centroids and
combines visual prototype similarity with cross-modal image-text
alignment for more reliable instance-level confidence estimation,
alongside task-adaptive thresholds calibrated automatically from
the training distribution.

\item We extend instance-aware Hard Gumbel gating symmetrically to
the text encoder, conditioned on image features, so that both
encoders co-adapt and co-suppress consistently for each input.
Extensive experiments on the MTIL benchmark demonstrate that CMAP
achieves state-of-the-art performance across all metrics while
introducing only 0.09M additional parameters over the baseline.

\end{itemize}


\section{Related Work}

\subsection{Incremental Learning}

Incremental learning requires a model to acquire new knowledge sequentially without
forgetting prior tasks, a challenge known as catastrophic forgetting~\cite{mccloskey1989catastrophic}.
Existing approaches fall into three families: regularization-based, rehearsal-based,
and architecture-based methods.

Regularization-based methods penalize changes to parameters important for previous tasks.
EWC~\cite{kirkpatrick2017overcoming} uses a Fisher information matrix to protect critical
weights, and MAS~\cite{aljundi2018memory} estimates weight importance from unlabeled data.
These methods scale poorly to large vision-language models such as CLIP, which contains
over 200M parameters, because meaningful protection of such a large parameter space requires
prohibitive computational and memory resources.

Rehearsal-based methods retain or reconstruct past knowledge during training.
LwF~\cite{li2017learning} uses knowledge distillation from the original model, and
iCaRL~\cite{rebuffi2017icarl} stores representative images per class.
LwF-VR~\cite{ding2022don} extends distillation to vision-language models but still
requires full fine-tuning.
Recent work has explored generative replay to avoid storing raw images, using
VLM-powered synthetic generation to rehearse old-task knowledge without explicit
memory buffers~\cite{mandalika2025replay}.
Despite this progress, replay-based approaches generally assume access to past data
or a capable generative model, which is impractical in privacy-sensitive or
annotation-constrained deployments.

Prompt-based parameter-efficient fine-tuning (PEFT) methods adapt frozen transformers
through a small set of learnable vectors, avoiding both full fine-tuning and explicit replay.
L2P~\cite{wang2022learning} and DualPrompt~\cite{wang2022dualprompt} use visual prompt-key
pools selected by cosine similarity, but assume unimodal, stable distributions that do not
hold across the visually diverse domains of MTIL.
S-Prompts~\cite{wang2023s} uses domain-specific prompts but requires task identity at
inference, which MTIL explicitly prohibits.
ZSCL~\cite{zheng2023preventing} preserves zero-shot transfer through teacher-student
distillation over external ImageNet wild data, but requires 211M trainable parameters and
out-of-domain data.
WiSE-FT~\cite{wortsman2022robust} interpolates between zero-shot and fine-tuned weights
as a lightweight alternative.

For multi-domain task-incremental learning, DIKI~\cite{tang2024mind} introduces
instance-aware prompting to handle domain shift without task identifiers, and
MoE-Adapter~\cite{yu2024boosting} applies mixture-of-experts adapters for domain
specialization at the cost of 59.8M parameters.
IAP~\cite{iap2025} improves upon DIKI with per-layer Hard Gumbel gating and a
two-stage Gaussian confidence cascade, reaching state-of-the-art performance with
only 2.4M parameters.
However, all existing MTIL methods control task routing and adaptation using visual
features alone, leaving CLIP's rich text embedding space entirely unexploited.

\subsection{Downstream Tasks of Vision-Language Models}

Contrastive Language-Image Pre-training (CLIP)~\cite{radford2021learning} learns a
joint visual-textual embedding space from large-scale image-text pairs, enabling strong
zero-shot generalization across diverse recognition tasks.
Adapting CLIP to downstream tasks through prompt learning has become a prominent
research direction.

CoOp~\cite{zhou2022learning} introduced learnable soft prompt vectors prepended to the
text encoder, outperforming manually designed templates in few-shot settings.
CoCoOp~\cite{zhou2022conditional} conditions text prompts on instance visual features,
improving generalization to classes unseen during training.
MaPLe~\cite{khattak2023maple} extends this idea to both the vision and language encoders
simultaneously through coupled multi-modal prompts, explicitly aligning the two modalities
during adaptation.

Recent work highlights the importance of cross-modal signal quality when training data is
scarce. Combining predictive prompts with negative learning has been shown to achieve
generalizable few-shot VLM adaptation~\cite{mandalika2025generalizable}, demonstrating
that leveraging joint visual-textual signals is especially critical under low-resource
conditions.

Existing approaches treat text embeddings as static class descriptors and condition
all adaptation signals on visual features alone, leaving CLIP's cross-modal text
embedding space unexploited for routing, confidence estimation, and encoder gating.


\begin{figure*}[t!]
\centering
\includegraphics[width=\textwidth]{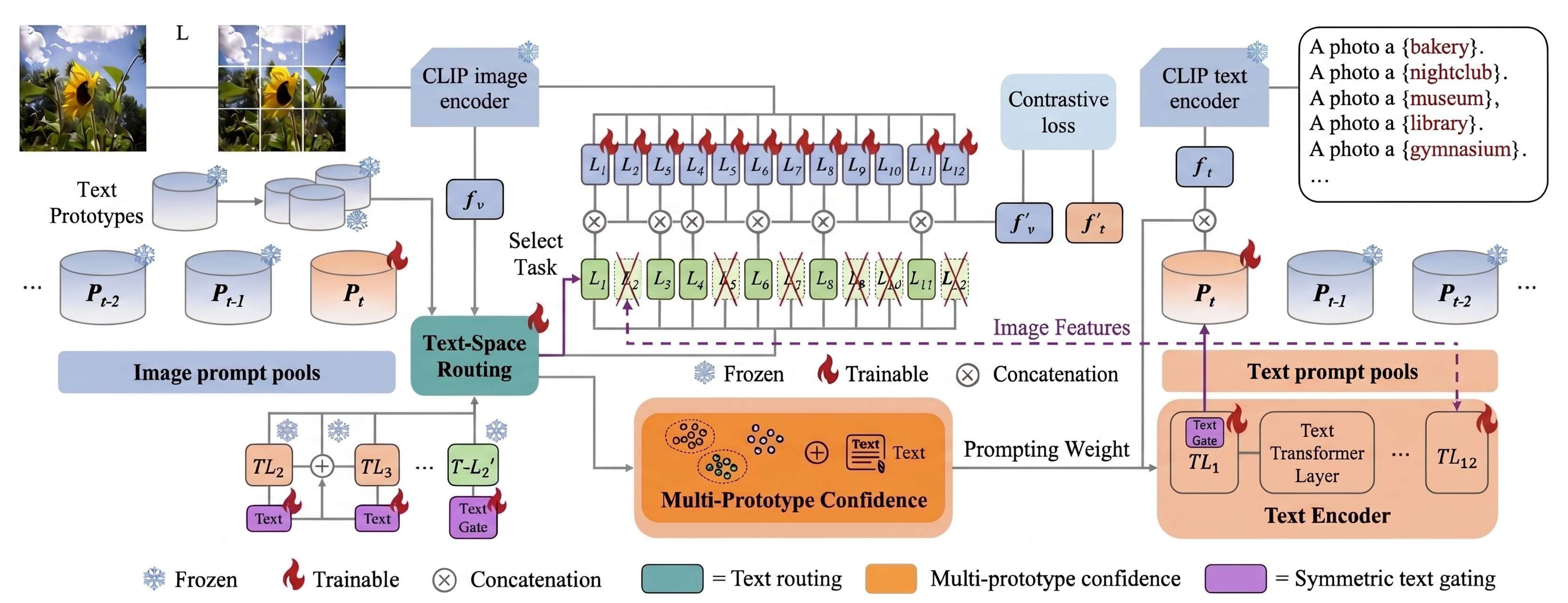}
\caption{
Overview of the proposed cross-modal adaptive prompting framework.
The input image is encoded by the frozen CLIP image encoder to produce visual features $f_v$, which are used by the Text-Space Routing module to select the nearest task prompt pool via cosine similarity to frozen text prototypes. Per-layer Hard Gumbel gates on the image encoder control prompt retrieval for each transformer layer, while the Multi-Prototype Confidence module combines K-means visual prototypes with cross-modal text alignment to produce a task-adaptive prompting weight.
On the text encoder side, symmetric gates conditioned on batch image features (purple dashed arrow) ensure both encoders adapt consistently, preserving cross-modal alignment on out-of-distribution inputs. Snowflake icons denote frozen components and flame icons denote trainable components.
\label{fig:architecture}
}
\end{figure*}

\section{Method}

\subsection{Preliminaries}

\noindent\textbf{MTIL Setup.}
A model sequentially learns $T$ tasks $\{\mathcal{D}_1, \dots, \mathcal{D}_T\}$, each with a
distinct visual domain and disjoint class set $\mathcal{Y}_t$.
At inference, no task identity is provided; the model must both identify the relevant
task and classify the input.
We adopt CLIP ViT-B/16 as the frozen backbone with embedding dimension $d{=}512$.
For a test image $\mathbf{x}$ and class name $c$, we define the frozen visual and
text features as:
\begin{equation}
    \mathbf{v} = f_v(\mathbf{x}) \in \mathbb{R}^d,
    \qquad
    \mathbf{e}_c = f_t\!\left(\text{``a photo of a } c\text{''}\right) \in \mathbb{R}^d,
    \label{eq:features}
\end{equation}
where $f_v$ and $f_t$ are the CLIP visual and text encoders.
We write $\text{sim}(\mathbf{a}, \mathbf{b}) = \mathbf{a}^\top \mathbf{b} / (\|\mathbf{a}\|\|\mathbf{b}\|)$
for cosine similarity throughout.
We follow the training protocol of~\cite{fu2026iap}, which maintains one learnable
prompt pool $\mathbf{P}_t \in \mathbb{R}^{l \times d}$ per task and applies
per-layer Hard Gumbel gating on the image encoder to control prompting intensity.
We adopt this setup as our baseline and address three gaps where visual-only adaptation
limits performance.

\subsection{Text-Space Task Routing}

Existing approaches select the task prompt by computing the log-probability of $\mathbf{v}$
under a per-task visual Gaussian fitted from training features~\cite{fu2026iap}.
This degrades when training data is scarce and when a new visual domain lies far from
any learned Gaussian, and it ignores the semantically stable class knowledge already
encoded in CLIP's text space.

We instead compute a mean text prototype for each task $t$ by averaging the text
embeddings of all class names in $\mathcal{Y}_t$:
\begin{equation}
    \boldsymbol{\tau}_t = \frac{1}{|\mathcal{Y}_t|} \sum_{c \in \mathcal{Y}_t} \mathbf{e}_c
    \in \mathbb{R}^d.
    \label{eq:prototype}
\end{equation}
Task selection at inference is then:
\begin{equation}
    t^* = \operatorname*{arg\,max}_{t \in \{1,\dots,T\}}\;
        \text{sim}\!\left(\mathbf{v},\, \boldsymbol{\tau}_t\right).
    \label{eq:routing}
\end{equation}
The prototypes $\{\boldsymbol{\tau}_t\}$ are computed once from frozen embeddings before
training begins and require no additional parameters.
Because they do not depend on the visual training distribution of any task, routing is
robust to task arrival order and degrades gracefully in the few-shot regime.

\subsection{Multi-Prototype Visual-Textual Confidence}

Prior work models each class as a single multivariate Gaussian over visual
features~\cite{fu2026iap}, which fits poorly when a class has a multi-modal visual
distribution (\eg aircraft photographed from different angles).
Fixed confidence thresholds ($\theta^{\text{up}}{=}0.8$, $\theta^{\text{low}}{=}0.2$)
are also applied uniformly across all tasks, ignoring task-specific distribution spread.

\noindent\textit{Multi-prototype representation.}
For each class $c$ in task $t$, we run K-means with $K{=}3$ over the visual training
features to obtain $K$ cluster centroids
$\{\mathbf{p}_{c,1}, \dots, \mathbf{p}_{c,K}\} \subset \mathbb{R}^d$.

\noindent\textit{Joint confidence score.}
Given test image $\mathbf{x}$ with feature $\mathbf{v}$, the visual and textual
confidence for class $c$ are:
\begin{align}
    \hat{C}_c^{\text{vis}}(\mathbf{x}) &=
        \max_{k=1}^{K}\; \text{sim}\!\left(\mathbf{v},\, \mathbf{p}_{c,k}\right),
    \label{eq:vis_conf} \\
    \hat{C}_c^{\text{txt}}(\mathbf{x}) &=
        \text{sim}\!\left(\mathbf{v},\, \mathbf{e}_c\right).
    \label{eq:txt_conf}
\end{align}
The joint confidence is their average:
\begin{equation}
    \hat{C}_c(\mathbf{x}) =
        \tfrac{1}{2}\,\hat{C}_c^{\text{vis}}(\mathbf{x})
        + \tfrac{1}{2}\,\hat{C}_c^{\text{txt}}(\mathbf{x}).
    \label{eq:joint_conf}
\end{equation}
The task-level confidence is the mean over the top-$k{=}5$ class scores:
\begin{equation}
    C_{t^*}(\mathbf{x}) = \frac{1}{k} \sum_{j=1}^{k} \hat{C}_{(j)}(\mathbf{x}),
    \label{eq:task_conf}
\end{equation}
where $\hat{C}_{(j)}(\mathbf{x})$ is the $j$-th largest value among
$\{\hat{C}_c(\mathbf{x})\}_{c \in \mathcal{Y}_{t^*}}$.

\noindent\textit{Task-adaptive thresholds.}
We calibrate $\theta_{t}^{\text{up}}$ and $\theta_{t}^{\text{low}}$ as the
80th and 20th percentiles of $\{C_{t}(\mathbf{x}_i)\}$ over the training set
of task $t$, requiring no manual tuning.
The prompting weight is:
\begin{equation}
    w(\mathbf{x}) =
    \begin{cases}
      1.0 & C_{t^*}(\mathbf{x}) > \theta_{t^*}^{\text{up}}, \\
      0.0 & C_{t^*}(\mathbf{x}) < \theta_{t^*}^{\text{low}}, \\
      C_{t^*}(\mathbf{x}) & \text{otherwise.}
    \end{cases}
    \label{eq:weight}
\end{equation}

\subsection{Symmetric Cross-Modal Gating}

In the adopted baseline, only the image encoder is gated.
When the image gate closes on an out-of-distribution (OOD) input, the image
representation reverts to zero-shot CLIP, but the text encoder still receives
task-specific prompts, disrupting cross-modal alignment.

We extend gating symmetrically to the text encoder.
Let $\bar{\mathbf{v}} = \frac{1}{B}\sum_{i=1}^{B} \mathbf{v}_i \in \mathbb{R}^d$
be the mean image feature over a batch of size $B$.
At each prompted text encoder layer $l$, a per-task per-layer linear projection
$\mathbf{W}_l \in \mathbb{R}^{2 \times d}$ produces a Hard Gumbel gate:
\begin{equation}
    g_l^{\text{txt}} =
        \operatorname{HardGumbel}\!\left(\mathbf{W}_l\,\bar{\mathbf{v}},\;
        \tau{=}3.0\right).
    \label{eq:txt_gate}
\end{equation}
The gated text layer output is:
\begin{equation}
    \mathbf{h}_l^{\text{txt}} =
        \mathbf{h}_{l-1}^{\text{txt}}
        + g_l^{\text{txt}} \cdot
        \operatorname{Attn}\!\left(\mathbf{h}_{l-1}^{\text{txt}},\, \mathbf{P}_l^{\text{txt}}\right),
    \label{eq:sym_gate}
\end{equation}
where $\mathbf{h}_{l-1}^{\text{txt}}$ is the text hidden state from the previous layer
and $\mathbf{P}_l^{\text{txt}}$ is the task prefix for layer $l$.
Because both image and text gates are conditioned on $\bar{\mathbf{v}}$, the two encoders
open and close together, preserving cross-modal alignment on OOD inputs.
The only additional parameters are $T \times L_{\text{txt}} \times 2d = 11 \times 8 \times
1{,}024 = 90{,}112$ weights (${\approx}0.09\text{M}$), bringing the total to ${\approx}2.5\text{M}$.

\subsection{Unified Inference Pipeline}

At test time, each image is routed to task $t^*$ via cosine similarity to frozen text
prototypes $\{\boldsymbol{\tau}_t\}$ (Eq.~\ref{eq:routing}). The joint confidence
$C_{t^*}(\mathbf{x})$ is computed over the classes of $t^*$ and mapped to a prompting
weight $w(\mathbf{x})$ through task-adaptive thresholds (Eq.~\ref{eq:weight}). The
image encoder runs with per-layer Hard Gumbel gating, while class templates for $t^*$
are encoded with symmetric text gating conditioned on $\bar{\mathbf{v}}$
(Eq.~\ref{eq:txt_gate}--\ref{eq:sym_gate}), keeping both encoders consistent with
batch-level distributional confidence. Classification is obtained via cosine similarity
between image and text features. Training follows a single-task protocol with no
cross-task replay.

\section{Experiments}

\begin{table*}[t!]
\centering
\fontsize{2.0}{2.4}\selectfont
\setlength{\tabcolsep}{0.2pt}
\renewcommand{\arraystretch}{0.42}
\caption{
Comparison with SOTA on MDCII benchmark in terms of ``Transfer'', ``Average'', and ``Last'' metrics (\%).
``XYZ'' denotes our method.
}
\label{tab:ord1}

\setlength{\tabcolsep}{4pt}
\renewcommand{\arraystretch}{1.05}

\resizebox{\textwidth}{!}{
\begin{tabular}{l l c c c c c c c c c c c c}

\toprule

\multirow{2}{*}{}
& \multirow{2}{*}{Method}
& \rotatebox{90}{Aircraft}
& \rotatebox{90}{Caltech101}
& \rotatebox{90}{CIFAR100}
& \rotatebox{90}{DTD}
& \rotatebox{90}{EuroSAT}
& \rotatebox{90}{Flowers}
& \rotatebox{90}{Food}
& \rotatebox{90}{MNIST}
& \rotatebox{90}{OxfordPet}
& \rotatebox{90}{Cars}
& \rotatebox{90}{SUN397}
& \multirow{2}{*}{Average} \\

&&&&&&&&&&&&& \\

\midrule

\multirow{2}{*}{CLIP}
& Zero-shot
& 24.3 & 88.4 & 68.2 & 44.6 & 54.9 & 71.0 & 88.5 & 59.4 & 89.0 & 64.7 & 65.2 & 65.3 \\

& Full Fine-tune
& 62.0 & 95.1 & 89.6 & 79.5 & 98.9 & 97.5 & 92.7 & 99.6 & 94.7 & 89.6 & 81.8 & 89.2 \\

\midrule

\multirow{12}{*}{\rotatebox{90}{\fontsize{3}{3}\selectfont\textbf{Transfer}}}

& Continual-FT
& -- & 67.1 & 46.0 & 32.1 & 35.6 & 35.0 & 57.7 & 44.1 & 60.8 & 20.5 & 46.6 & 44.6 \\

& LwF
& -- & 74.5 & 56.9 & 39.1 & 51.1 & 52.6 & 72.8 & 60.6 & 75.1 & 30.3 & 55.9 & 58.9 \\

& iCaRL
& -- & 56.6 & 44.6 & 32.7 & 39.3 & 46.6 & 68.0 & 46.0 & 77.4 & 31.9 & 60.5 & 50.4 \\

& LwF-VR
& -- & 77.1 & 61.0 & 40.5 & 45.3 & 54.4 & 74.6 & 47.9 & 76.7 & 36.3 & 58.6 & 57.2 \\

& WISE-FT
& -- & 73.5 & 55.6 & 35.6 & 41.5 & 47.0 & 68.3 & 53.9 & 69.3 & 26.8 & 51.9 & 52.3 \\

& ZSCL
& -- & 86.0 & 67.4 & \underline{45.4} & \underline{50.4}
& 69.1 & 87.6 & 61.8 & 86.8 & 60.1 & \underline{66.8} & 68.1 \\

& L2P
& -- & 65.6 & 50.9 & 30.4 & 41.4 & 49.3 & 71.8 & 36.3 & 77.5 & 55.3 & 53.4 & 53.2 \\

& DualPrompt
& -- & 56.7 & 51.4 & 28.7 & 33.7 & 45.6 & 70.9 & 59.5 & 77.7 & 49.5 & 50.4 & 52.4 \\

& S-Prompt
& -- & 67.3 & 49.4 & 26.7 & 39.7 & 47.1 & 70.2 & 34.3 & 78.9 & 56.7 & 52.2 & 52.2 \\

& MoE-Adapter
& -- & 87.9 & 68.2 & 44.4 & 49.9 & \underline{70.7}
& \underline{88.7} & 59.7 & 89.1 & 64.5 & 65.5 & 68.9 \\

& DIKI
& -- & 92.9 & \underline{69.1} & 43.2 & 43.9
& 65.4 & 85.3 & 56.0 & 88.4 & 64.0 & 65.6 & 67.4 \\

& IAP
& -- & \underline{93.0} & 68.7 & 44.0 & 47.0
& 70.4 & 85.9 & \underline{63.5} & \underline{89.7}
& \underline{66.2} & 63.3 & \underline{69.2} \\

& XYZ (Ours)
& -- & \textbf{94.3} & \textbf{71.9} & \textbf{47.5} & \textbf{54.2}
& \textbf{73.9} & \textbf{91.3} & \textbf{67.7} & \textbf{94.0}
& \textbf{70.4} & \textbf{71.1} & \textbf{73.6} \\

\midrule

\multirow{12}{*}{\rotatebox{90}{\fontsize{3}{3}\selectfont\textbf{Average}}}

& Continual-FT
& 25.5 & 81.5 & 59.1 & 53.2 & 64.7 & 51.8 & 63.2 & 64.3 & 69.7 & 31.8 & 49.7 & 55.9 \\

& LwF
& 36.3 & 86.9 & 72.0 & 59.0 & 73.7 & 60.0 & 73.6 & 74.8 & 80.0 & 37.3 & 58.1 & 64.7 \\

& iCaRL
& 35.5 & 89.2 & 72.2 & 60.6 & 68.8 & 70.0 & 78.2 & 62.3 & 81.8 & 41.2 & 62.5 & 65.7 \\

& LwF-VR
& 29.6 & 87.7 & 74.4 & 59.5 & 72.4 & 63.6 & 77.0 & 66.7 & 81.2 & 43.7 & 60.7 & 65.1 \\

& WISE-FT
& 26.7 & 86.5 & 64.3 & 57.1 & 65.7 & 58.7 & 71.1 & 70.5 & 75.8 & 36.9 & 54.6 & 60.7 \\

& ZSCL
& 45.1 & 92.0 & 80.1 & 64.3 & 79.5 & 81.6
& \underline{89.6} & 75.2 & 88.9 & 64.7 & \underline{68.0} & 75.4 \\

& L2P
& 38.0 & 85.2 & 78.2 & 61.3 & 72.9 & 74.9 & 79.7 & 59.1 & 82.0 & 59.7 & 55.4 & 67.9 \\

& DualPrompt
& 37.8 & 84.3 & 78.6 & 60.1 & 71.1 & 73.2 & 79.1 & 73.9 & 82.3 & 55.1 & 52.8 & 68.0 \\

& S-Prompts
& 37.5 & 92.5 & 77.5 & 58.2 & 76.4 & 74.1 & 78.8 & 57.9 & 83.0 & 60.8 & 54.4 & 68.3 \\

& MoE-Adapter
& \underline{50.2} & 91.9 & 83.1 & 69.4 & 78.9 & 84.0 & 89.1 & 73.7 & 89.3 & 67.7 & 66.9 & 76.7 \\

& DIKI
& 45.4 & \underline{95.7} & 83.0 & 65.0 & 78.2 & 82.5 & 87.1 & 71.7 & 90.0 & 67.2 & 66.6 & 75.7 \\

& IAP
& 45.9 & 95.8 & \underline{83.3}
& \underline{66.5} & \underline{79.5} & \underline{84.8}
& 87.5 & \underline{76.6} & \underline{91.0}
& \underline{69.2} & 64.5 & \underline{76.8} \\

& XYZ (Ours)
& \textbf{55.9} & \textbf{96.2} & \textbf{87.5}
& \textbf{70.0} & \textbf{83.7} & \textbf{88.1}
& \textbf{93.8} & \textbf{80.6} & \textbf{95.2}
& \textbf{72.8} & \textbf{71.8} & \textbf{81.4} \\

\midrule

\multirow{12}{*}{\rotatebox{90}{\fontsize{3}{3}\selectfont\textbf{Last}}}

& Continual-FT
& 31.0 & 89.3 & 65.8 & 67.3 & 88.9 & 71.1 & 85.6 & 99.6 & 92.9 & 77.3 & 81.1 & 77.3 \\

& LwF
& 26.3 & 87.5 & 71.9 & 66.6 & 79.9 & 66.9 & 83.8 & 99.6 & 92.1 & 66.1 & 80.4 & 74.6 \\

& iCaRL
& 35.8 & 93.0 & 77.0 & 70.2 & 83.3 & 88.5 & 90.4 & 86.7 & 93.2 & 81.2 & \textbf{81.9} & 80.1 \\

& LwF-VR
& 20.5 & 89.8 & 72.3 & 67.6 & 85.5 & 73.8 & 85.7 & 99.6 & 93.1 & 73.3 & 80.9 & 76.6 \\

& WISE-FT
& 27.2 & 90.8 & 68.0 & 68.9 & 86.9 & 74.0 & 87.6 & 99.6 & 92.6 & 77.8 & 81.3 & 77.7 \\

& ZSCL
& 40.6 & 92.2 & 81.3 & 70.5 & 94.8 & 90.5 & 91.9 & 98.7 & 93.9 & \underline{85.3} & 80.2 & 83.6 \\

& L2P
& 38.0 & 87.1 & 84.2 & 72.9 & 86.0 & 96.1 & 89.2 & 99.0 & 94.1 & 79.6 & 76.0 & 82.0 \\

& DualPrompt
& 37.8 & 87.1 & 84.6 & 71.8 & 89.2 & 96.3 & 89.1 & 99.1 & 94.5 & 79.9 & 76.5 & 82.3 \\

& S-Prompt
& 37.5 & 95.1 & 83.7 & 70.2 & 97.5 & 96.5 & 89.0 & 99.1 & 94.0 & 79.5 & 75.8 & 83.4 \\

& MoE-Adapter
& \underline{49.8} & 92.2 & 86.1 & \underline{78.1}
& 95.7 & 94.3 & 89.5 & 98.1 & 89.9 & 81.6 & 80.0 & 85.0 \\

& DIKI
& 45.4 & 95.9 & 86.0 & 73.0 & 97.8 & 96.8 & 89.3 & 99.3 & 94.4 & 81.8 & 76.4 & 85.1 \\

& IAP
& 46.8 & \underline{96.1} & \underline{86.7}
& 75.2 & \underline{98.1} & \underline{97.0}
& \underline{89.6} & \underline{99.4} & \underline{94.7}
& 82.8 & 76.7 & \underline{85.7} \\

& XYZ (Ours)
& \textbf{51.0} & \textbf{96.7} & \textbf{92.4}
& \textbf{85.0} & \textbf{98.4} & \textbf{98.2}
& \textbf{93.3} & \textbf{99.6} & \textbf{96.0}
& \textbf{85.5} & \underline{80.2} & \textbf{88.9} \\

\bottomrule
\end{tabular}
}
\end{table*}

\begin{table}[t!]
\centering
\caption{
    Ablation study on Order-I (\%).
    Each row removes one component from the full model.
}
\label{tab:ablation1}
\begin{tabular}{lccc}
\toprule
Method & Transfer & Average & Last \\
\midrule
w/o text routing  & 71.3 & 79.1 & 88.4 \\
w/o MPVTC         & 72.8 & 79.2 & 87.9 \\
w/o sym.\ gating  & 71.5 & 78.5 & 87.1 \\
\midrule
\textbf{XYZ (full)} & \textbf{73.6} & \textbf{81.4} & \textbf{88.9} \\
\bottomrule
\end{tabular}
\end{table}

\begin{table}[t!]
\centering
\caption{
    Effect of task routing strategy on Order-I (\%).
}
\label{tab:ablation_routing}
\begin{tabular}{lccc}
\toprule
Routing & Transfer & Average & Last \\
\midrule
Visual Gaussian       & 69.2 & 76.8 & 85.7 \\
Visual mean prototype & 71.8 & 79.3 & 87.4 \\
\textbf{Text prototype (ours)} & \textbf{73.6} & \textbf{81.4} & \textbf{88.9} \\
\bottomrule
\end{tabular}
\end{table}


\begin{table*}[t!]
\centering

\fontsize{3.4}{3.6}\selectfont
\setlength{\tabcolsep}{0.2pt}
\renewcommand{\arraystretch}{0.42}

\caption{
Comparison with SOTA on MDCII benchmark in terms of ``Transfer'', ``Average'', and ``Last'' metrics (\%).
``XYZ'' denotes our method.
}

\label{tab:ord2}

\setlength{\tabcolsep}{4pt}
\renewcommand{\arraystretch}{1.05}

\resizebox{\textwidth}{!}{
\begin{tabular}{l l c c c c c c c c c c c c}

\toprule

&
\multirow{2}{*}{Method}
& \rotatebox{90}{StanfordCars}
& \rotatebox{90}{Food}
& \rotatebox{90}{MNIST}
& \rotatebox{90}{OxfordPet}
& \rotatebox{90}{Flowers}
& \rotatebox{90}{SUN397}
& \rotatebox{90}{Aircraft}
& \rotatebox{90}{Caltech101}
& \rotatebox{90}{DTD}
& \rotatebox{90}{EuroSAT}
& \rotatebox{90}{CIFAR100}
& \multirow{2}{*}{Average}
\\

&&&&&&&&&&&&& \\

\midrule

\multirow{2}{*}{CLIP}

& Zero-shot
& 64.7 & 88.5 & 59.4 & 89.0 & 71.0 & 65.2 & 24.3 & 88.4 & 44.6 & 54.9 & 68.2 & 65.3 \\

& Full Fine-tune
& 89.6 & 92.7 & 99.6 & 94.7 & 97.5 & 81.8 & 62.0 & 95.1 & 79.5 & 98.9 & 89.2 & 89.2 \\

\midrule

\multirow{12}{*}{\rotatebox{90}{\fontsize{4}{4}\selectfont\textbf{Transfer}}}

& Continual-FT
& & \textbf{89.5} & \underline{59.6} & 57.9 & 40.0 & 46.7 & 11.1 & 70.0 & 30.5 & 26.6 & 37.7 & 46.6 \\

& LwF
& & 87.8 & 58.5 & 71.9 & 46.6 & 57.3 & 12.8 & 81.4 & 34.5 & 34.5 & 46.8 & 53.2 \\

& iCaRL
& & 86.1 & 51.8 & 67.6 & 50.4 & 57.9 & 11.0 & 72.3 & 31.2 & 32.7 & 48.1 & 50.9 \\

& LwF-VR
& & 88.2 & 57.0 & 71.4 & 50.0 & 58.0 & 13.0 & 82.0 & 34.4 & 29.3 & 47.6 & 53.1 \\

& WISE-FT
& & 87.2 & 57.6 & 67.0 & 45.1 & 54.0 & 12.9 & 78.6 & 35.5 & 28.4 & 44.3 & 51.1 \\

& ZSCL
& & 88.3 & 57.5 & 84.7 & 68.1 & \underline{64.8} & 21.1 & 88.2 & \underline{45.3} & \textbf{55.2} & \underline{68.2} & 64.1 \\

& L2P
& & 70.6 & 30.7 & 78.3 & 42.8 & 38.3 & 17.4 & 75.3 & 27.4 & 23.1 & 20.7 & 42.5 \\

& DualPrompt
& & 79.9 & 46.9 & 85.2 & 51.3 & 45.1 & 9.3 & 82.7 & 29.9 & 42.9 & 47.2 & 52.1 \\

& S-Prompts
& & 59.8 & 46.2 & 67.7 & 47.5 & 43.8 & 13.5 & 76.8 & 31.4 & 22.6 & 43.5 & 45.3 \\

& MoE-Adapter
& & 88.8 & 59.5 & \underline{89.1} & 69.9 & 64.4 & 18.1 & 86.9 & 43.7 & 54.6 & 68.2 & 64.3 \\

& DIKI
& & 85.8 & 55.3 & 89.5 & 71.1 & 62.9 & 23.7 & 93.6 & 42.1 & 43.4 & 67.9 & 63.5 \\

& IAP
& & 85.7 & 59.4 & \underline{89.1} & \underline{71.3} & 62.7 & \underline{24.4} & \underline{94.0} & 43.8 & 49.0 & \underline{68.6} & \underline{64.9} \\

& XYZ (Ours)
& & \underline{89.0} & \textbf{61.5} & \textbf{92.3} & \textbf{75.8} & \textbf{66.4} & \textbf{31.0} & \textbf{95.1} & \textbf{53.7} & \underline{55.1} & \textbf{70.0} & \textbf{69.9} \\

\midrule

\multirow{12}{*}{\rotatebox{90}{\fontsize{4}{4}\selectfont\textbf{Average}}}

& Continual-FT
& 42.1 & 70.5 & 92.2 & 80.1 & 54.5 & 59.1 & 19.8 & 78.3 & 41.0 & 38.1 & 42.3 & 56.2 \\

& LwF
& 49.0 & 77.0 & 92.1 & 85.9 & 66.5 & 67.2 & 20.9 & 84.7 & 44.6 & 45.5 & 50.5 & 62.2 \\

& iCaRL
& 52.0 & 75.9 & 77.4 & 74.6 & 58.4 & 59.3 & 11.7 & 79.6 & 42.1 & 43.2 & 51.7 & 56.9 \\

& LwF-VR
& 44.9 & 75.8 & 91.8 & 85.3 & 63.5 & 67.6 & 16.9 & 84.9 & 44.0 & 40.6 & 51.3 & 60.6 \\

& WISE-FT
& 52.6 & 79.3 & 91.9 & 83.9 & 63.4 & 65.2 & 23.3 & 83.7 & 45.4 & 40.0 & 48.2 & 61.5 \\

& ZSCL
& 81.7 & \textbf{91.3} & 91.1 & 91.0 & 82.9 & \underline{72.5} & 33.6 & 89.7 & \underline{53.3} & \underline{62.8} & 69.9 & 74.5 \\

& L2P
& 80.1 & 87.4 & 86.7 & 89.6 & 76.8 & 59.1 & 27.7 & 79.5 & 33.9 & 34.6 & 26.5 & 62.5 \\

& DualPrompt
& 78.6 & 88.4 & 89.7 & 91.7 & 80.0 & 62.4 & 23.2 & 85.0 & 41.3 & 51.6 & 50.7 & 67.5 \\

& S-Prompts
& 79.2 & 86.5 & 89.5 & 87.0 & 78.2 & 61.5 & 25.5 & 83.6 & 41.9 & 36.3 & 47.2 & 65.1 \\

& MoE-Adapter
& \underline{84.9} & 89.9 & 89.3 & 91.4 & 86.2 & 72.2 & 33.4 & 89.4 & 53.3 & 61.4 & 69.9 & 74.7 \\

& DIKI
& 84.8 & 89.0 & 91.3 & 93.2 & 87.8 & 72.2 & 34.0 & \underline{94.5} & 50.9 & 53.3 & 69.6 & 74.2 \\

& IAP
& 82.5 & \underline{89.2} & \textbf{99.4} & \underline{94.9} & 88.0 & 70.4 & 34.3 & 94.4 & 52.3 & 57.9 & \underline{70.2} & 75.1 \\

& XYZ (Ours)
& \textbf{86.8} & 88.6 & \underline{97.2} & \textbf{95.1} & \textbf{90.7} & \textbf{72.9} & \textbf{37.5} & \textbf{96.1} & \textbf{57.0} & \textbf{65.1} & \textbf{72.6} & \textbf{78.1} \\

\midrule

\multirow{12}{*}{\rotatebox{90}{\fontsize{4}{4}\selectfont\textbf{Last}}}

& Continual-FT
& 24.0 & 67.3 & 99.1 & 87.4 & 44.3 & 67.0 & 29.5 & 92.3 & 61.3 & 81.0 & 88.1 & 67.4 \\

& LwF
& 34.6 & 69.6 & 99.3 & 88.7 & 61.1 & 72.5 & 32.5 & 88.1 & 65.6 & 90.9 & 87.9 & 71.9 \\

& iCaRL
& 46.0 & 81.5 & 91.3 & 82.8 & 66.5 & 72.2 & 16.3 & 91.6 & 68.1 & 83.2 & 87.8 & 71.6 \\

& LwF-VR
& 27.4 & 61.2 & 99.4 & 86.3 & 60.6 & 70.7 & 23.4 & 88.0 & 61.3 & 84.3 & 88.1 & 68.2 \\

& WISE-FT
& 35.6 & 76.9 & \textbf{99.5} & 89.1 & 62.1 & 71.8 & 27.8 & 90.8 & 67.0 & 85.6 & \underline{87.6} & 72.2 \\

& ZSCL
& 78.2 & \textbf{91.1} & 97.6 & 92.5 & 87.4 & \underline{78.2} & 45.0 & 92.3 & 72.7 & 96.2 & 86.3 & 83.4 \\

& L2P
& 80.1 & 89.1 & 99.1 & 93.8 & 96.2 & 76.5 & 40.1 & 86.9 & 73.5 & 86.3 & 84.2 & 82.3 \\

& DualPrompt
& 78.6 & 88.3 & 99.2 & 94.1 & 96.5 & 76.8 & 39.8 & 89.0 & 71.6 & 90.7 & 84.9 & 82.8 \\

& S-Prompts
& 79.2 & 88.1 & 99.1 & 94.3 & 95.8 & 76.3 & 39.9 & 95.5 & 70.1 & 97.6 & 84.4 & 83.8 \\

& MoE-Adapter
& \underline{84.1} & 88.5 & 94.0 & 91.8 & 94.1 & 77.8 & 50.4 & 93.3 & \underline{77.1} & 87.7 & 86.6 & 84.1 \\

& DIKI
& 81.8 & 88.3 & \underline{99.3} & \underline{94.7} & \underline{97.4} & 76.8 & \underline{46.4} & 96.0 & 74.2 & \textbf{98.0} & 86.0 & 85.4 \\

& IAP
& 82.5 & 88.6 & \textbf{99.4} & \textbf{94.9} & \textbf{97.7} & 76.9 & 46.1 & \textbf{96.1} & 74.7 & \textbf{98.0} & 86.6 & \underline{85.9} \\

& XYZ (Ours)
& \textbf{86.8} & \underline{90.2} & 99.0 & \textbf{94.9} & 97.2 & \textbf{82.5} & \textbf{50.4} & \underline{95.7} & \textbf{77.9} & \underline{97.6} & \textbf{88.3} & \textbf{87.4} \\

\bottomrule
\end{tabular}
}
\end{table*}

\begin{table}[t!]
\centering
\caption{
    Effect of confidence combination in MPVTC on Order-I (\%).
}
\label{tab:ablation_conf}
\begin{tabular}{lccc}
\toprule
Confidence & Transfer & Average & Last \\
\midrule
Visual only   & 72.6 & 80.2 & 88.2 \\
Textual only  & 72.2 & 79.8 & 88.1 \\
\textbf{Equal weight (ours)} & \textbf{73.6} & \textbf{81.4} & \textbf{88.9} \\
\bottomrule
\end{tabular}
\end{table}

\begin{table}[t!]
\centering
\caption{
    Comparison of trainable parameters and memory buffer requirements.
    \checkmark/\texttimes{} denotes buffer required or not required.
    Results are reported on Order-I.
}
\label{tab:params}
\resizebox{\columnwidth}{!}{
\begin{tabular}{lcccccc}
\toprule
Method & Buffer & Params & Transfer & Average & Last \\
\midrule
iCaRL~\cite{rebuffi2017icarl}     & \checkmark & 211M  & 50.4 & 65.7 & 80.1 \\
ZSCL~\cite{zheng2023preventing}   & \checkmark & 211M  & 68.1 & 75.4 & 83.6 \\
MoE-Adapter~\cite{yu2024boosting} & \texttimes & 59.8M & 68.9 & 76.7 & 85.0 \\
DIKI~\cite{tang2024mind}          & \texttimes & 1.8M  & 67.4 & 75.7 & 85.1 \\
IAP~\cite{fu2026iap}                & \texttimes & 2.4M  & 69.2 & 76.8 & 85.7 \\
\midrule
\textbf{XYZ (ours)}               & \texttimes & \textbf{2.5M} & \textbf{73.6} & \textbf{81.4} & \textbf{88.9} \\
\bottomrule
\end{tabular}
}
\end{table}

\begin{table*}[t!]
\centering
\vspace{-2mm}

\fontsize{3.4}{3.6}\selectfont
\setlength{\tabcolsep}{0.2pt}
\renewcommand{\arraystretch}{0.42}

\caption{
Comparison with SOTA on 16-shot MTIL benchmark in terms of ``Transfer'', ``Average'', and ``Last'' metrics (\%).
``XYZ'' denotes our method. We label the best and second methods with bold and underline styles.
}

\label{tab:16shots}

\setlength{\tabcolsep}{4pt}
\renewcommand{\arraystretch}{1.05}

\resizebox{\textwidth}{!}{
\begin{tabular}{l l c c c c c c c c c c c c}
\toprule

\multirow{2}{*}{Method}
& \rotatebox{90}{Aircraft}
& \rotatebox{90}{Caltech101}
& \rotatebox{90}{CIFAR100}
& \rotatebox{90}{DTD}
& \rotatebox{90}{Flowers}
& \rotatebox{90}{Food}
& \rotatebox{90}{StanfordCars}
& \rotatebox{90}{SUN397}
& \multirow{2}{*}{Average}
\\

\midrule

\multirow{2}{*}{CLIP}

& Zero-shot
& 24.8 & 92.9 & 68.4 & 43.8 & 71.4 & 85.8 & 65.8 & 62.6 & 64.4 \\

& Full Fine-tune
& 62.0 & 96.2 & 89.6 & 79.5 & 97.5 & 92.7 & 89.6 & 81.8 & 86.1 \\

\midrule

\multirow{6}{*}{\rotatebox{90}{\fontsize{2.4}{2.4}\selectfont\textbf{Transfer}}}

& ZSCL
& -- & 87.3 & 67.7 & 45.4 & 67.8 & 86.6 & 59.7 & 63.4 & 68.3 \\

& L2P
& -- & 66.7 & 54.3 & 30.6 & 47.3 & 71.5 & 54.6 & 52.4 & 53.9 \\

& DualPrompt
& -- & 78.8 & 64.4 & 32.0 & 51.7 & 77.5 & 49.4 & 51.3 & 57.9 \\

& S-Prompts
& -- & 70.3 & 52.7 & 31.5 & 54.8 & 74.0 & 55.4 & 50.0 & 55.5 \\

& DIKI
& -- & 92.7 & 68.8 & 44.1 & 70.0 & 86.2 & 65.1 & 65.5 & 70.3 \\

& IAP
& -- & \underline{93.2} & \underline{68.9} & \underline{44.5} & \underline{71.4} & \underline{85.5} & \underline{66.1} & \underline{65.4} & \underline{70.9} \\

& XYZ (Ours)
& -- & \textbf{96.1} & \textbf{76.3} & \textbf{52.4} & \textbf{78.0} & \textbf{91.1} & \textbf{74.0} & \textbf{72.7} & \textbf{77.1} \\

\midrule

\multirow{6}{*}{\rotatebox{90}{\fontsize{2.4}{2.4}\selectfont\textbf{Average}}}

& ZSCL
& 33.5 & 90.5 & 74.7 & \underline{58.5} & 79.7 & 87.7 & 64.8 & 64.8 & 69.3 \\

& L2P
& 30.2 & 84.5 & 70.1 & 51.9 & 69.6 & 77.1 & 60.0 & 55.2 & 62.3 \\

& DualPrompt
& 36.5 & 89.5 & 72.5 & 52.7 & 72.3 & 80.8 & 56.1 & 54.2 & 64.3 \\

& S-Prompts
& 30.6 & 86.8 & 70.0 & 51.7 & 74.3 & 78.5 & 60.7 & 53.0 & 63.2 \\

& DIKI
& 41.3 & \underline{95.3} & 76.5 & 58.5 & 82.2 & 86.4 & 68.2 & \underline{66.6} & 71.9 \\

& IAP
& \textbf{42.5} & 94.8 & \underline{77.6} & \underline{59.3} & \underline{82.5} & \underline{86.5} & \underline{69.6} & 65.4 & \underline{72.5} \\

& XYZ (Ours)
& \textbf{46.7} & \textbf{95.3} & \textbf{80.9} & \textbf{63.7} & \textbf{85.9} & \textbf{89.7} & \textbf{73.5} & \textbf{70.6} & \textbf{75.9} \\

\midrule

\multirow{6}{*}{\rotatebox{90}{\fontsize{2.4}{2.4}\selectfont\textbf{Last}}}

& ZSCL
& 27.7 & 90.9 & 74.4 & 64.7 & 90.2 & \underline{89.2} & \underline{80.6} & 74.6 & 74.0 \\

& L2P
& 30.2 & 87.1 & 75.4 & 64.7 & 91.9 & 86.4 & 76.1 & 74.7 & 73.3 \\

& DualPrompt
& 36.5 & 91.0 & 75.1 & 65.1 & 92.9 & 86.2 & 76.2 & 74.2 & 74.7 \\

& S-Prompts
& 30.6 & 89.2 & 75.8 & 63.8 & 93.9 & 86.2 & 76.7 & 73.9 & 73.8 \\

& DIKI
& 41.3 & 95.6 & 79.0 & 67.3 & 94.4 & 86.8 & 77.6 & 74.4 & 77.1 \\

& IAP
& \underline{42.5} & \underline{95.8} & \underline{78.6} & \underline{68.1} & \underline{95.3} & 87.5 & 79.2 & \underline{74.8} & \underline{77.7} \\

& XYZ (Ours)
& \textbf{46.7} & \textbf{95.8} & \textbf{81.2} & \textbf{71.4} & \textbf{96.8} & \textbf{93.4} & \textbf{82.0} & \textbf{77.9} & \textbf{80.7} \\

\bottomrule

\end{tabular}
}

\end{table*}

\noindent\textbf{Baselines and Implementation.}
We compare against two groups of methods: full-parameter fine-tuning approaches
(Continual-FT, LwF~\cite{li2017learning}, iCaRL~\cite{rebuffi2017icarl},
LwF-VR~\cite{ding2022don}, WiSE-FT~\cite{wortsman2022robust},
ZSCL~\cite{zheng2023preventing}) and parameter-efficient methods
(L2P~\cite{wang2022learning}, DualPrompt~\cite{wang2022dualprompt},
S-Prompts~\cite{wang2023s}, MoE-Adapter~\cite{yu2024boosting},
DIKI~\cite{tang2024mind}), with IAP~\cite{iap2025} as the direct baseline.
All methods use a frozen CLIP ViT-B/16 backbone~\cite{radford2021learning}.
Training follows IAP with SGD (lr~5.0, 10~epochs per task), prompt depth~12
for the image encoder and~8 for the text encoder, prompt length~8, Gumbel
temperature~3.0, and batch size~128.

\noindent\textbf{Datasets and Metrics.}
We evaluate on the MTIL benchmark~\cite{tang2024mind}, comprising 11 visual-recognition
datasets spanning 1{,}201 classes: Aircraft, Caltech101, CIFAR-100, DTD, EuroSAT,
Flowers102, Food101, MNIST, OxfordPet, StanfordCars, and SUN397.
Following~\shortcite{zheng2023preventing}, we report Transfer (forward generalisation),
Last (retention on seen tasks), and Average (integrating both), under two task
orderings and a 16-shot data-scarce setting.

\subsection{Main Results}

Table~\ref{tab:ord1} and Table~\ref{tab:ord2} report results under Order-I and
Order-II respectively.
Under Order-I, Transfer reaches 73.6\%, Average 81.4\%, and Last 88.9\%, while
under Order-II the corresponding figures are 69.9\%, 78.1\%, and 87.4\%.
The Transfer metric shows the largest absolute gains in both orderings, consistent
with the expectation that text-space routing is grounded in frozen CLIP embeddings
rather than accumulated visual statistics and therefore generalises more reliably
to tasks not yet encountered during training.
The Transfer gap over the visual-Gaussian baseline holds at 4.4\% under Order-I
and 5.0\% under Order-II, indicating that routing via frozen text prototypes is
order-independent by construction.
On a per-dataset basis, the largest gains concentrate in visually fine-grained or
texture-rich domains such as DTD, Aircraft, and EuroSAT where unimodal visual
Gaussians struggle most, while performance on compact distributions like MNIST
and Caltech101 remains strong across all methods.

Table~\ref{tab:16shots} reports results in the data-efficient 16-shot setting
across eight datasets.
The Transfer gap over the visual-Gaussian baseline widens to 6.2\% relative to
4.4\% in the full-data regime, which is structurally expected as text prototypes
are derived entirely from frozen CLIP embeddings and are independent of the number
of visual training samples, whereas visual distributional statistics degrade as
sample count decreases.
The Average and Last metrics follow a similar trend, with the proposed method
reaching 75.9\% and 80.7\% respectively, confirming that the multi-prototype
confidence module retains its discriminative quality even when only 16 training
samples per task are available.

\subsection{Ablation Study}

\noindent\textbf{Component-wise ablation.}
Table~\ref{tab:ablation1} isolates each component by removing it from the full model.
Removing text-space routing produces the largest Transfer drop to 71.3\%, confirming
its role in forward generalisation with negligible effect on Last (88.4\%).
Removing MPVTC reduces Average most uniformly to 79.2\%, reflecting its contribution
to sustained performance across the task sequence.
Removing symmetric gating causes the largest Average and Last drops to 78.5\% and
87.1\%, identifying cross-modal coherence as the principal driver of knowledge retention.

\noindent\textbf{Effect of confidence combination.}
Table~\ref{tab:ablation_conf} compares using visual prototypes alone, text embeddings
alone, and the equal-weight combination within MPVTC.
Visual-only and textual-only configurations reach 72.6\% and 72.2\% Transfer
respectively, both below the combined 73.6\%, confirming that the two modalities
capture complementary information and neither alone is sufficient for reliable
confidence estimation across visually diverse domains.

\noindent\textbf{Effect of routing strategy.}
Table~\ref{tab:ablation_routing} compares three routing strategies.
Replacing visual Gaussian routing with a visual mean prototype raises Transfer from
69.2\% to 71.8\%, showing that the Gaussian distributional assumption is itself a
source of error.
Switching to text prototypes further raises Transfer to 73.6\%, a gain attributable
specifically to the cross-modal signal in CLIP text embeddings rather than the change
in similarity metric alone.

\noindent\textbf{Parameter efficiency.}
Table~\ref{tab:params} compares trainable parameter counts and memory buffer requirements.
The proposed method uses 2.5M parameters, a 0.1M increase over the visual-Gaussian
baseline from text encoder gating weights alone, with no memory buffer required.
Despite this, it outperforms MoE-Adapter (59.8M) and full fine-tuning methods (211M)
by substantial margins, using 24$\times$ and 84$\times$ fewer parameters respectively.

\section{Conclusion}

This paper identified three cross-modal gaps in existing Multi-Domain Task-Incremental
Learning methods and addressed each with a targeted contribution: text-space task
routing via frozen CLIP text prototypes, multi-prototype joint visual-textual
confidence estimation with task-adaptive thresholds, and symmetric Hard Gumbel gating
extended to the text encoder conditioned on batch image features.
The proposed method reaches 73.6\% Transfer, 81.4\% Average, and 88.9\% Last under
Order-I with consistent margins under Order-II, and the Transfer gain widens to 6.2\%
in the 16-shot setting, confirming that text prototypes scale gracefully where visual
distributional statistics degrade.
All three contributions together require approximately 2.5M trainable parameters with
no external data or memory buffer, demonstrating that exploiting CLIP's pre-trained
cross-modal alignment is a more parameter-efficient adaptation strategy than scaling
model capacity, which directly aligns with the goal of generalising effectively from
limited resources.

\appendix

\section*{Ethical Statement}
Generative AI tools were used solely to improve language clarity and readability. All technical content, experimental results, and scientific claims remain the authors' own responsibility.

\bibliographystyle{named}
\bibliography{ijcai26}

\end{document}